\pdfoutput=1

\documentclass[11pt]{article}

\usepackage[preprint]{coling}

\usepackage{times}
\usepackage{latexsym}
\usepackage{amsmath}

\usepackage[T1]{fontenc}

\usepackage[utf8]{inputenc}

\usepackage{microtype}

\usepackage{inconsolata}

\usepackage{graphicx}
\usepackage{amsmath}
\usepackage{algorithm}
\usepackage{algpseudocode}
\usepackage{booktabs}
\usepackage{subcaption}
\usepackage{wrapfig}
\usepackage{float}

\title{Cross-lingual transfer of multilingual models on low resource African Languages}

\author{
Harish~Thangaraj $^{*,1}$, Ananya~Chenat $^{*,1}$, 
Jaskaran~Singh~Walia$^{1}$ and Vukosi~Marivate$^{2}$\\    
$^{1}$School of Computer Science and Engineering, Vellore Institute of Technology, Chennai, India\\
$^{2}$Department of Computer Science, University of Pretoria\\
{\small $^{*}$ Email: ananyachenat483@gmail.com}
}

\begin{document}
\maketitle
\begin{abstract}
Large multilingual models have significantly advanced natural language processing (NLP) research. However, their high resource demands and potential biases from diverse data sources have raised concerns about their effectiveness across low-resource languages. In contrast, monolingual models, trained on a single language, may better capture the nuances of the target language, potentially providing more accurate results. This study benchmarks the cross-lingual transfer capabilities from a high-resource language to a low-resource language for both, monolingual and multilingual models, focusing on Kinyarwanda and Kirundi, two Bantu languages. We evaluate the performance of transformer based architectures like Multilingual BERT (mBERT), AfriBERT, and BantuBERTa against neural based architectures such as BiGRU, CNN, and char-CNN. The models were trained on Kinyarwanda and tested on Kirundi, with fine-tuning applied to assess the extent of performance improvement and catastrophic forgetting. AfriBERT achieved the highest cross-lingual accuracy of 88.3\% after fine-tuning, while BiGRU emerged as the best-performing neural model with 83.3\% accuracy. We also analyze the degree of forgetting in the original language post-fine-tuning. While  monolingual models remain competitive, this study highlights that multilingual models offer strong cross-lingual transfer capabilities in resource limited settings. 
\end{abstract}
\section{Introduction}
Recent advancements in natural language processing (NLP) have led to the development of both monolingual and multilingual models, with substantial progress in high-resource languages. However, low-resource languages continue to face significant challenges due to limited data and corpus availability, which restrict the development and performance of language models. Cross-lingual transfer learning, where knowledge from a resource-rich language is transferred to a lexically similar low-resource language, has emerged as a promising solution to this problem.

The multilingual architectures like multilingual BERT (mBERT) \cite{BERT} are trained on a variety of languages. This broad training pattern allows them to generalize and recognize patterns across several languages. Yet on the downside, these models are highly influenced by the dataset used. A biased training set inclined towards larger corpus from a certain language can potentially lead to sub-optimal performance on  underrepresented languages. Monolingual models, on the other hand, are trained exclusively on a single language, allowing them to capture finer linguistic details and nuances. 

 To analyze the performance of these types, this work studies the transfer from Kinyarwanda to Kirundi (Bantu family) using both monolingual and multilingual models. Instances of Multilingual BERT (mBERT) \cite{BERT}, AfriBERT \cite{AfriBERT}and BantuBERTa \cite{parvess2024bantuberta,parvess2023thesis}, are tested for the multilingual scenario. Convolutional Neural Networks (CNN), Character-Level Convolutional Neural Networks (char-CNN), and Bi-Directional Gated Re-current Units (BiGRU) are evaluated for the monolingual scenario \cite{2}. The models are trained on Kinyarwanda and then tested and benchmarked on Kirundi before and after fine tuning. We also estimate the extent of catastrophic forgetting of the models on the initial language after fine tuning. We test our initial hypothesis of monolingual models outperforming multilingual models considering the linguistic similarity between the two languages and the ability to capture intrinsic nuances is tested. While existing research focuses on neural and multilingual models separately, this study provides a comprehensive comparison which will aid future scholars to use the findings. 
\section{Related work}
\subsection{NLP for low resource languages}
Low resource languages (LRL) have gained increasing attention by researchers in recent years with the growth of Natural Language Processing tasks. Limited corpora, fewer linguistic tools, and a lack of digital resources have posed the need for research techniques to mitigate these challenges. \cite{14} review past and future techniques such as transfer learning, data augmentation, sentence level alignment and multilingual embeddings providing general trends in processing LRL and giving an overview of techniques available for our study. Data augmentation for LRL are explored by \cite{13} in their study using Assamese and isiZulu promising potential improvement in model performance in low resource settings. \cite{12} outline neural machine translation between a high resource and low resource language by effectively back-translating monolingual LRL data to create an enhanced corpus. Their study provides a compelling technique to handle data limitations of LRL with structurally similar high resource language data.

\subsection{Transfer learning}
Cross-lingual transfer emerges as a powerful and practical approach to model resource limited languages without abundant availability of linguistically similar secondary language data. Utilization of existing resources for learning transfer offers faster convergence and multilingual downstream capability on the two or more languages.  \cite{4} analyse methods for sentiment classification for LRLs by 
introducing annotation project and direct transfer as two transfer learning approaches using partial lexicalization and LSTM architecture. Results indicated that single-source transfer from English generally outperformed the baseline for all languages. The  direct transfer approach opens a promising avenue when the source and target languages are from the same family as in our case. \cite{11} propose UniBridge , an adapter based architecture incorporating embedding initialization and multi-source transfer. The experiment results in substantial performance improvement especially owing to the embedding initialization which allows better adaptation to low resource languages.  \cite{2} explore NLP for Kirundi, focusing on multiclass classification using cross-lingual transfer from Kinyarwanda. Two new datasets, KINNEWS and KIRNEWS, were introduced, along with stop word lists for both languages. For cross-lingual text classification, Kinyarwanda embeddings were used to train models, which were then tested on the Kirundi corpus, leveraging the mutual intelligibility of the languages. Results showed that BiGRU performed best on KINNEWS, while CNN excelled on KIRNEWS in cross-lingual settings, suggesting that BiGRU requires a larger dataset for optimal performance, presenting a compelling base paper for this work. 

\subsection{Monolingual models for transfer}
Monolingual models focus on a single language, leveraging language-specific features and resources to achieve higher accuracy for tasks like translation, text generation, and classification. By training solely on one language, these models can better capture linguistic nuances.
 \cite{15} explore cross linugal transfer of monolingual models. The study ulitlizes BERT models from various languages and fine tuned using the GLUE benchmark. The researchers study two probing techniques namely, structural probing that evaluates how the embeddings capture syntactic structures and semantic probing to determine if words are used with the same meaning in different contexts. The probing results indicated that knowledge from the source language enhanced the learning of both syntactic and semantic aspects in the target language.
The research by \cite{16}. examines cross-lingual representation learning by introducing a method that transfers monolingual models to other languages without requiring shared subword vocabularies or joint pre-training along with the introduction of the XQuaD dataset. The methodology involves pre-training an English model and then learning new subword embeddings for other languages. The findings suggest that monolingual representations effectively generalize across languages. 
 \cite{8} present the Bilingual Document Representation Learning model (BiDRL) learning document representations using a joint learning algorithm to capture both semantic and sentiment correlations between bilingual texts using a shared embedding space. BiDRL significantly outperformed state-of-the-art methods across nine tasks involving English (source language) and Japanese, German, and French (target languages) achieving an accuracy of 81.34\%. \cite{5} test cross-multilingual transfer for Moroccan sentiment analysis, focusing on Arabic specifc models and a monolingual model (DarijaBERT) using training and validation datasets.  Among the models, DarijaBERT, despite being trained on a smaller scale of data, outperformed most of the multilingual models, demonstrating the effectiveness of monolingual models for specific dialects. 

\subsection{Multilingual models for transfer}
Multilingual models facilitate cross-lingual transfer by creating shared linguistic representations across different languages, enabling knowledge transfer from well-resourced languages to those with fewer resources. This approach is a promising area of research, offering significant potential for advancing NLP in underrepresented languages.
\cite{parvess2023thesis,parvess2024bantuberta} evaluates the state of current multilingual models and explores the potential of the Bantu language family due to its topographical similarity. The study introduces BantuBERTa, a multilingual model primarily trained on low-resourced, topographically similar languages, and benchmarks it against AfriBERT, mBERT, and XLM-R. Results revealed that although BantuBERTa had relatively lower scores compared to other models, indicated successful generalization between Bantu languages with an F1 score greater than 50\%. \cite{1} aims to develop a universal model for cross-lingual text classification in low-resource languages. \mbox{IndicSBERT} and LaBSE models were trained on samples from Tamil, Malayalam, Marathi, Oriya, and Telugu, and tested on Bengali, Kannada, Gujarati, and Punjabi. Results demonstrated that IndicSBERT generally outperforms LaBSE, showcasing strong multilingual and cross-lingual capabilities. \cite{6} evaluate the Multi-View Encoder-Classifier (MVEC) model against various models like multilingual BERT (mBERT) and XLM for cross-lingual sentiment classification. MVEC outperformed these models in 8 out of 11 sentiment classification tasks across five language pairs, employing unsupervised machine translation and language discriminator to align latent space between languages.  \cite{7} introduce XLM-R, a large-scale multilingual language model trained on 100 languages using two terabytes of CommonCrawl data. XLM-R offers better performance than models such as mBERT, particularly in low-resource languages such as Swahili and Urdu. The study also highlights increasing the model's capacity helps mitigate capacity challenges as the languages increase. \cite{9} introduce ARBERT and MARBERT, two deep bidirectional transformer-based models designed for Arabic language processing, focusing on Modern Standard Arabic (MSA) and various dialects. Results demonstrated that ARBERT and MARBERT achieved new state-of-the-art performance, with MARBERT excelling in social media-related tasks due to its extensive training on dialectal data. \cite{10} present a comparative analysis of task-specific pre-training and cross-lingual transfer techniques for sentiment analysis in Dravidian code-switched languages, specifically Tamil-English and Malayalam-English.  The experiments demonstrate that task-specific pre-training consistently outperforms cross-lingual transfer in both zero-shot and supervised settings. The study also explores the potential of combining cross-lingual transfer with task-specific pre-training by fine-tuning TweetEval on the Hinglish dataset before adapting it to Tamil-English and Malayalam-English.

\subsection{Modelling for African languages}
Modelling for African languages has gained increasing attention due to the need for inclusive natural language processing (NLP) systems. These languages, often underrepresented in global datasets, present unique challenges such as limited resources, diverse linguistic structures, and dialectal variations. Recent advancements, including the development of multilingual models and language-specific datasets, have made significant strides in addressing these issues.
 \cite{18} explore the performance of various language models, including n-gram, AWD-LSTM, QRNN, and transformer architectures, specifically within the context of South African languages. Their results indicate that the AWD-LSTM and QRNN consistently outperform other models, such as n-gram and Basic-LSTM, across multiple datasets, achieving better bits-per-character metrics. Furthermore, the study highlights the advantages of multilingual training, where incorporating data from related languages significantly enhances model performance for isiZulu and Sepedi. The research presented by  \cite{19}  explores multilingual neural machine translation (NMT) strategies for African languages. The findings highlight that while traditional single-pair NMT models (S-NMT) exhibit limitations, more advanced methodologies such as semi-supervised NMT (SS-NMT) and transfer learning (TL) significantly enhance performance, particularly in out-of-domain settings. Notably, the multilingual model (M-NMT) consistently outperformed S-NMT in multiple translation directions, achieving particularly striking improvements for the least-resourced language pairs. The papers \cite{20, 17} investigate the effectiveness of multilingual language models pretrained on low-resource African languages, specifically Amharic, Hausa, and Swahili. The study reveals that multilingual models generally outperform monolingual ones in transfer effectiveness and emphasize the necessity for pre-training methods.
\section{Experiments}
We benchmark and evaluate the performance of the monolingual and multilingual models on distinct datasets (from the same language family) by following a training pipeline to train these models with their best hyperparameters.


\subsection{Dataset}

This study employs 2 distinct datasets, one in Kinyarwanda and the other in Kirundi sourced from \cite{2}.

\begin{table}[H]\centering
\begin{tabular}{@{}lp{6cm}@{}}\toprule
\textbf{Field} & \textbf{Description} \\ \midrule
label & Numerical labels ranging from 1 to 14 \\
en\_label & English labels \\
kin\_label & Kinyarwanda labels \\
kir\_label & Kirundi labels \\
url & The link to the news source \\
title & The title of the news article \\
content & The full content of the news article \\ \bottomrule
\end{tabular}
\caption{Field descriptions of the raw dataset}
\label{table:1}
\end{table}

\begin{table}[H]\centering
\begin{tabular}{@{}lp{6cm}@{}}\toprule
\textbf{Field} & \textbf{Description} \\ \midrule
label & Numerical labels ranging from 1 to 14 \\
title & The title of the news article \\
content & The full content of the news article \\ \bottomrule
\end{tabular}
\caption{Field descriptions of the cleaned dataset}
\label{table:2}
\end{table}

For the Kinyarwanda dataset, news articles from various websites and newspapers were used. A total of 21268 articles are distributed across 14 classes, with a train:test split ratio as 17014:4254. Similarly for the Kirundi dataset, a total of 4612 articles are distributed across 14 classes, with a train:test split ratio as 3690:922. For both Kinyarwada and Kirundi, the cleaned versions of the datasets were taken from the codebase affiliated with the research paper \cite{2}, which thereafter served as the primary reference for our data preprocessing steps. Each dataset contains 3 main fields: 1) Label, which comprises of numerical labels ranging from 1 to 14 representing the category of the article, 2) Title, which is the title of the news article and 3) Content, the full content of the news article as summarised in Table \ref{table:1} and Table \ref{table:2}.

\subsection{Transformer Models}

\begin{algorithm}[H]
\caption{BERT Tuning: Tokenization, Training on Kinyarwanda, Fine-tuning on Kirundi, and Cross-Lingual Evaluation}\label{alg:bert_training}
\begin{algorithmic}[1]
\Statex \textbf{Inputs:}
\Statex BERT base model $M$
\Statex Kinyarwanda corpus $\mathcal{D}_{Kinyarwanda}$
\Statex Kirundi corpus $\mathcal{D}_{Kirundi}$

\Statex \textbf{Output:}
\Statex Evaluated cross-lingual metrics $\mathcal{E}$

\State \textbf{Pre-training Phase}
\State Tokenize $\mathcal{D}_{Kinyarwanda}$
\State Train $M$ on tokenized $\mathcal{D}_{Kinyarwanda}$ for the specified task (News classification)
\State Save trained model as $M_{trained}$

\State \textbf{Fine-tuning Phase}
\State Load $M_{trained}$
\State Tokenize $\mathcal{D}_{Kirundi}$
\State Fine-tune $M_{trained}$ on tokenized $\mathcal{D}_{Kirundi}$ for the downstream task
\State Save fine-tuned model as $M_{finetuned}$

\State \textbf{Evaluation Phase}
\State Evaluate $M_{finetuned}$ on cross-lingual benchmarks
\State Compute cross-lingual metrics $\mathcal{E}$ (accuracy, F1-score)
\State \textbf{Return} $\mathcal{E}$
\end{algorithmic}
\end{algorithm}
We explore three large pre-trained architectures : Multilingual BERT (mBERT) \cite{BERT}, AfriBERT \cite{AfriBERT} and BantuBERTa \cite{parvess2023thesis,parvess2024bantuberta} on the pipeline given by Algorithm \ref{alg:bert_training}. Being an extension of the original BERT model with a pre-trained corpus of Wikipedia data from 104 different high resource and low resource languages, mBERT promises effective cross-lingual transfer learning use cases aided by its shared representations across languages. AfriBERT, trained on a diverse corpus of 11 African languages is designed to address the unique linguistic characteristics and challenges of African languages including Kinyarwanda and Kirundi, the two languages tested in this work encouraging favourable transfer on the downstream task. Bantu languages share certain linguistic features, and BantuBERTa leverages these commonalities to enhance performance in Natural Language Processing within the language family. As opposed to mBERT and AfriBERT, BantuBERTa is pre-trained on a smaller dataset which puts its transfer abilities and accuracy to test within this experiment.

\subsection{Neural Models}

\begin{algorithm}[H]
\caption{Monolingual Neural Model Training: Embeddings, Fine-tuning, and Cross-Lingual Evaluation}\label{alg:general_monolingual_training}
\begin{algorithmic}[1]
\Statex \textbf{Inputs:}
\Statex Neural model $M$ (CNN, BiGRU, etc.)
\Statex Kinyarwanda corpus $\mathcal{D}_{Kinyarwanda}$
\Statex Kinyarwanda embeddings $\mathcal{E}_{Kinyarwanda}$
\Statex Kirundi corpus $\mathcal{D}_{Kirundi}$
\Statex Kirundi embeddings $\mathcal{E}_{Kirundi}$

\Statex \textbf{Output:}
\Statex Evaluated cross-lingual metrics $\mathcal{E}$

\State \textbf{Pre-training Phase}
\State Load pre-trained Kinyarwanda embeddings $\mathcal{E}_{Kinyarwanda}$
\State Tokenize Kinyarwanda corpus $\mathcal{D}_{Kinyarwanda}$
\State Map tokenized $\mathcal{D}_{Kinyarwanda}$ to $\mathcal{E}_{Kinyarwanda}$
\State Train $M$ on $\mathcal{E}_{Kinyarwanda}$ for the specified task (e.g., News classification)
\State Save trained model as $M_{trained}$

\State \textbf{Fine-tuning Phase}
\State Load $M_{trained}$
\State Tokenize Kirundi corpus $\mathcal{D}_{Kirundi}$
\State Map tokenized $\mathcal{D}_{Kirundi}$ to $\mathcal{E}_{Kirundi}$
\State Fine-tune $M_{trained}$ on $\mathcal{E}_{Kirundi}$ for the downstream task
\State Save fine-tuned model as $M_{finetuned}$

\State \textbf{Evaluation Phase}
\State Evaluate $M_{finetuned}$ on cross-lingual benchmarks
\State Compute cross-lingual metrics $\mathcal{E}$ (accuracy, F1-score, etc.)
\State \textbf{Return} $\mathcal{E}$
\end{algorithmic}
\end{algorithm}

\setlength{\belowcaptionskip}{10pt}
\begin{figure*}[h!]
\centering
\includegraphics[width=16cm]{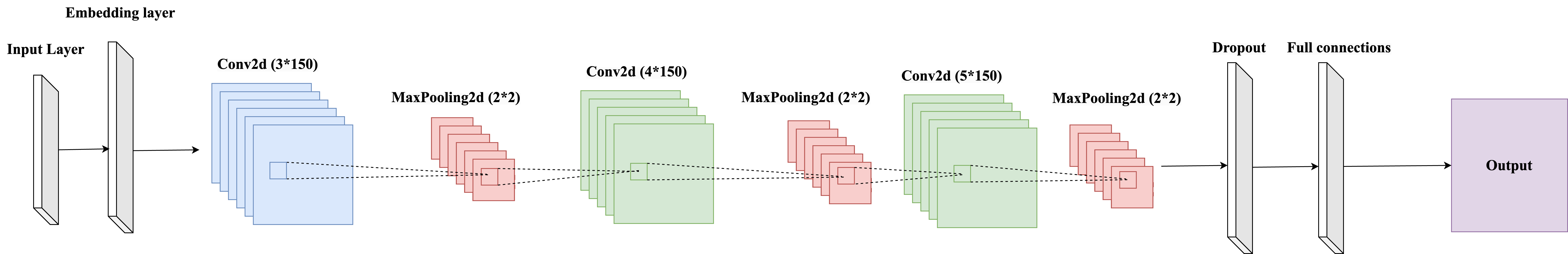}
\caption{Architecture for CNN implementation}
\label{fig:cnn}
\end{figure*}

\setlength{\belowcaptionskip}{10pt}
\begin{figure*}[h!]
\centering
\includegraphics[width=16cm]{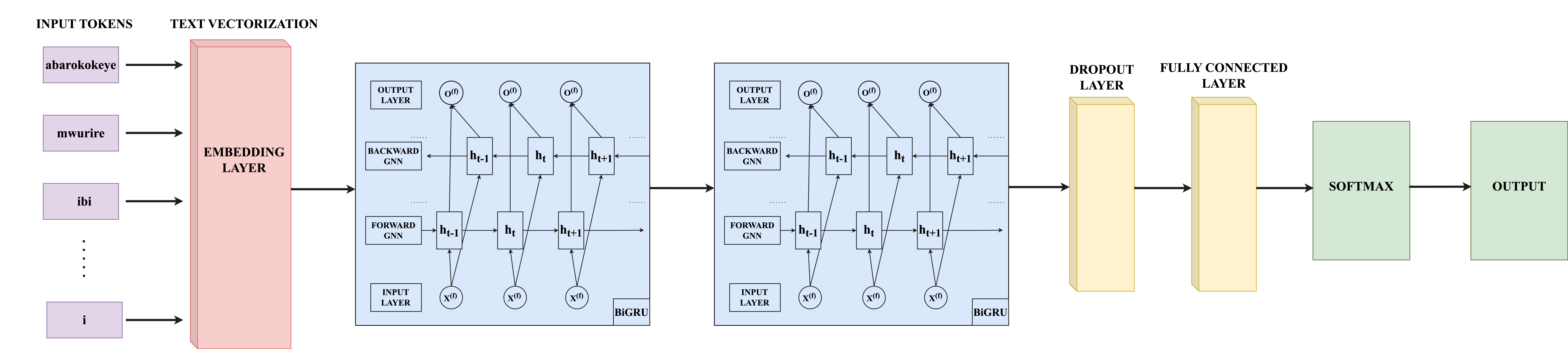}
\caption{Architecture for BiGRU implementation}
\label{fig:bigru}
\end{figure*}

Convolutional Neural Networks (CNN), Character-Level Convolutional Neural Networks (char-CNN), and Bi-Directional Gated Recurrent Units (BiGRU) are the three neural models evaluated for cross-lingual transfer in this study \cite{2} as described by Algorithm\ref{alg:general_monolingual_training}. While CNNs are widely recognized for their image processing capabilities, they also perform effectively in language tasks by treating text as a sequential input and applying filters to extract essential features across multiple layers as represented in Figure \ref{fig:cnn}. Char-CNNs build on the CNN framework by focusing on characters rather than whole words, applying convolutional filters to individual characters. This approach is particularly useful for languages with complex morphological structures, as it allows the model to capture subtle linguistic details, enhancing transfer performance. BiGRU as in \ref{fig:bigru}, is a recurrent neural network (RNN) designed to process sequential text data in both forward and backward directions, thereby capturing more comprehensive contextual information, which is crucial for effective cross-lingual transfer.

\subsection{Learning Scenario}

For a unified text representation, title and content fields were merged into a single field labeled ‘text’. With a vector size of 50, window size of 5 and word frequency threshold of 5, a Word2Vec model was trained adopting a skip-gram model and hierarchical Softmax to obtain word embeddings. The labels were converted the zero based for easier model training and classification capabilities. 
The three BERT architectures were loaded along with its tokenizers and trained initially on the Kinyarwanda dataset preparing it for the downstream task classification of 14 labels. The initial learning involved training for 8, 25 and 8 epochs for the models mBERT, AfriBERT and BantuBERTa respectively, with a batch size of 32. 500 warmup steps were employed to stabilize training, while a weight decay of 0.01 was applied to prevent overfitting. The model was evaluated based on steps with a log interval of 10. Given the computation on a Mac environment, MPS was opted due the unavailability of CUDA for enhanced performance over CPU training, refer Table \ref{tab:bert_parameters}. Metrics were first evaluated on the Kinyarwanda test dataset to ensure that the models had effectively learned the language-specific features and performed well on the source language. The cross-lingual transfer was tested in three steps. One being the direct transfer, where the Kinywarnda trained model was directly applied on Kirundi to benchmark results, without fine tuning on Kirundi. The second step being post fine-tuning transfer, where the Kinyarwanda trained model was fine tuned on the Kirundi dataset after which evaluation was done. Lastly, Evaluating on Kinyarwanda again after fine tuning to understand the extent of forgetting the initial language calculated as percentage.

\begin{table}[h]\centering
\caption{Parameters of BERT models}
\begin{tabular}{@{}ll@{}}\toprule
\textbf{Parameter} & \textbf{Value} \\ \midrule
\textbf{Number of Labels} & 14 \\
\textbf{Input Sequence Length} & 128 \\
\textbf{Truncation} & True \\
\textbf{Padding} & True \\
\textbf{Device} & \texttt{MPS} \\
\textbf{Number of Training Epochs:} \\
    mBERT & 8 \\
    AfriBERT & 25 \\
    BantuBERTa & 8 \\
\textbf{Training Batch Size} & 32 \\
\textbf{Evaluation Batch Size} & 32 \\
\textbf{Warmup Steps} & 500 \\
\textbf{Weight Decay} & 0.01 \\
\textbf{Logging Steps} & 10 \\
\textbf{Load Best Model at End} & True \\
\textbf{Evaluation Strategy} & \texttt{steps} \\ \bottomrule
\end{tabular}
\label{tab:bert_parameters}
\end{table}

For the neural models, the preprocessed training dataset was loaded and divided into 90\% training and 10\% validation sets. The Natural Language Toolkit (NLTK) tokenizer was applied to the text corpus, followed by building the vocabulary using custom-trained Kinyarwanda embeddings with a vector size of 50. Initially, the models were trained on Kinyarwanda and evaluated on the corresponding test set to assess intra-language learning. Subsequently, the trained model was evaluated on Kirundi both directly and after fine-tuning. Finally, the fine-tuned model was tested again on Kinyarwanda to examine any potential forgetting.

\subsection{Evaluation Metrics}
To comprehensively assess performance of the various architectures and compare it with overall performance we utilize a set of evaluation metrics that cover various aspects of effectiveness in all the modalities.

\textbf{Average Accuracy (\%):} This metric measures the test-set accuracy across the downstream task at the end of the learning process. It is calculated as:

\textbf{F1 Score:} For tasks involving multiple classification, we use the F1 score, which balances precision and recall, offering a more nuanced view of the model's performance.
Where \textit{Precision} is the ratio of correctly predicted positive observations to the total predicted positives and  \textit{Recall} is the ratio of correctly predicted positive observations to all observations in the actual class.

\textbf{Average Forgetting:} This metric measures the average reduction in performance for previously learned tasks when new tasks are introduced. It quantifies how much the model forgets prior knowledge as it learns new information. Average forgetting can be calculated as the mean difference between the maximum accuracy achieved for each task and the final accuracy after all tasks have been learned.

\begin{equation}
\small
\begin{split}
\text{Forgetting (\%)} &= \\
& \left( \frac{\text{Performance}_{\text{before}} - \text{Performance}_{\text{after}}}{\text{Performance}_{\text{before}}} \times 100 \right)
\end{split}
\end{equation}

These evaluation metrics are utilized to assess the transfer performances, across all the varying models.

\section{Results}

\renewcommand{\arraystretch}{1}
\begin{table*}[h]\centering
\caption{Metrics describing cross-lingual testing on Kirundi}
\begin{tabular}{@{}lcccc@{}}\toprule
Model & Accuracy before FT & F1 before FT & Accuracy after FT & F1 after FT \\ \midrule
mBERT & 0.5872 & 0.5917 & 0.8462 & 0.8422 \\
Afri BERT & 0.7421 & 0.7474 & \textbf{0.8830} & \textbf{0.8787} \\
Bantu BERT & 0.7454 & 0.7375 & 0.8657 & 0.8606 \\
BiGRU & 0.2404 & 0.2300 & \textbf{0.8332} & \textbf{0.8790} \\
CNN & 0.2190 & 0.2320 & 0.5913 & 0.5732 \\
Char-CNN & 0.1916 & 0.1621 & 0.4879 & 0.4764 \\ \bottomrule
\end{tabular}
\label{cross-lingual}
\end{table*}

\renewcommand{\arraystretch}{1}
\begin{table*}[h]\centering
\caption{Comparison of metrics testing on Kinyarwanda}
\begin{minipage}{.45\linewidth}
\centering
\subcaption{Performance on Kinyarwanda before fine tuning\\}
\vspace{5pt}
\begin{tabular}{@{}lcc@{}}\toprule
Model & Accuracy & F1 score \\ \midrule
mBERT & 0.7884 & 0.7747 \\
Afri BERT & 0.8498 & 0.8447 \\
Bantu BERT & 0.8601 & 0.8555 \\
BiGRU & \textbf{0.8851} & 0.8434 \\
CNN & 0.8740 & \textbf{0.8660} \\
Char-CNN & 0.6930 & 0.6823 \\ \bottomrule
\end{tabular}
\label{tab:initial_kinyarwanda}
\end{minipage}
\quad
\begin{minipage}{.45\linewidth}
\centering
\subcaption{Performance \& Forgetting post fine tuning\\}
\vspace{5pt}
\begin{tabular}{@{}lcc@{}}\toprule
Model & Accuracy & Forget \% \\ \midrule
mBERT & 0.7645 & \textbf{3.03} \\
Afri BERT & \textbf{0.8061} & 5.14 \\
Bantu BERT & 0.2172 & 74.00 \\
BiGRU & 0.2329 & 73.68 \\
CNN & 0.2207 & 74.86 \\
Char-CNN & 0.1968 & 71.50 \\ \bottomrule
\end{tabular}
\label{tab:forget_kinyarwanda}
\end{minipage}
\end{table*}



The results depicted by Table \ref{cross-lingual} show that Afribert outperforms mBERT and BantuBERT in the tested transfer scenario. Post fine-tuning, AfriBERT attained the highest accuracy of 88.3\% on the Kirundi test set suggesting its strong capability in learning the target language. The mBERT and BantuBERT models performed competitively, attaining an accuracy of 84.6\% and 86.5\% post fine-tuning on Kirundi. AfriBERT and BantuBERT produced better metrics than mBERT during the initial testing on Kinywarnda proving their better suitability for African languages (refer Table \ref{tab:initial_kinyarwanda}).  When re-evaluated on the Kinyarwanda dataset after fine tuning, AfriBERT and mBERT produced minimal forgetting of 5.14\% and 3.03\% favouring their cross-lingual transfer use-cases. On the contrary, BantuBERT suffered from catastophic forgetting whose implications are discussed under limitations (refer Table \ref{tab:forget_kinyarwanda}).

Among all the neural models, the metrics in Table \ref{cross-lingual} show BiGRU emerging as a strong choice, attaining an accuracy of 83.3\% on Kirundi after fine-tuning. CNN and Char-CNN both offer average performance in the transfer with 59.1\% and 48.7\% accuracy scores respectively. All three architectures undergo catastrophic forgetting as given by Table \ref{tab:forget_kinyarwanda} when evaluated on Kinywarnda post fine-tuning. Regardless, BiGRU and CNN present compelling metrics when trained and tested on Kinyarwanda directly proving its monolingual capabilities.

\setlength{\belowcaptionskip}{10pt}
\begin{figure}[h!]
\centering
\includegraphics[width=8cm]{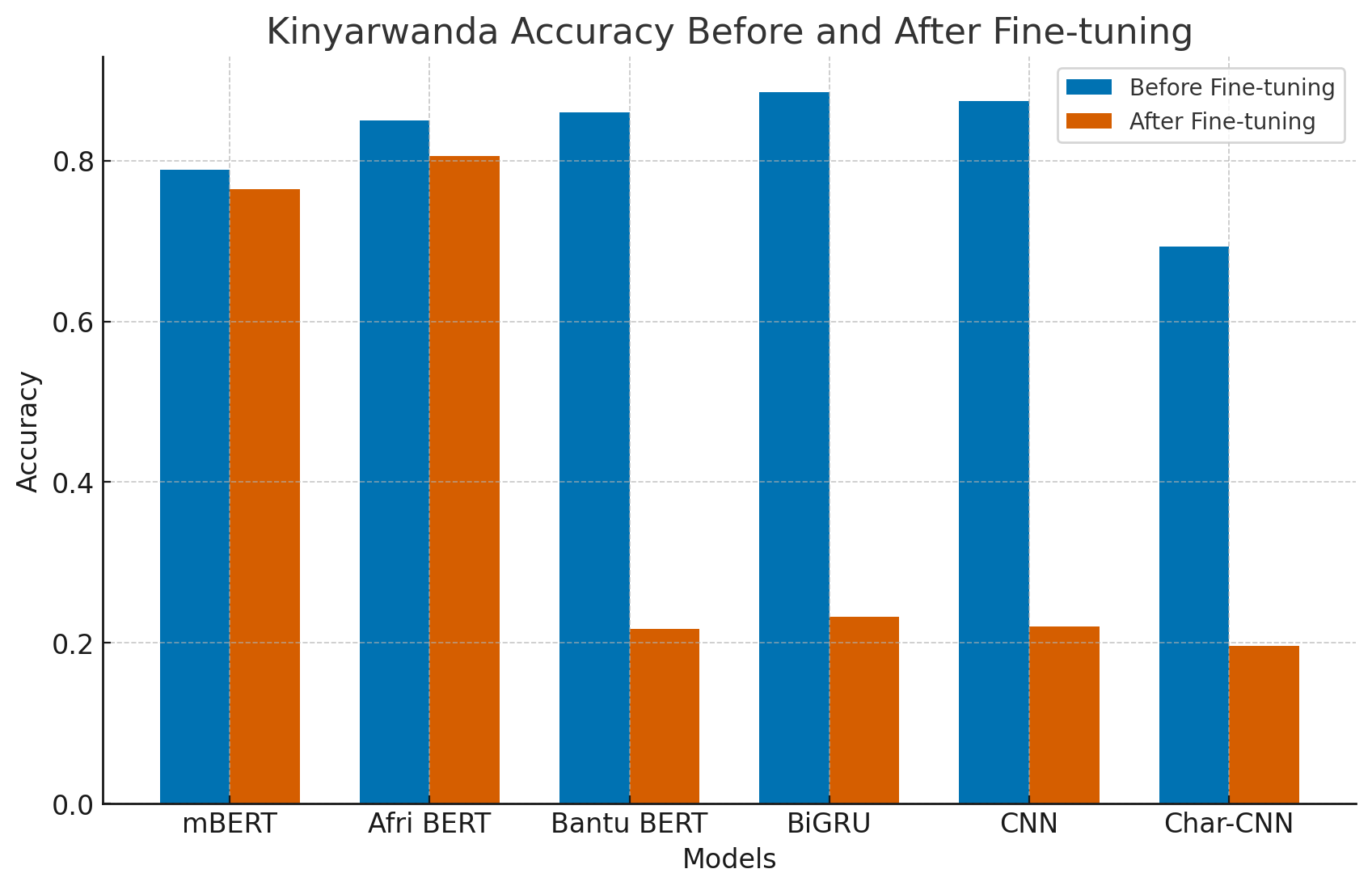}
\caption{Performance (Accuracy and F1) on Kinyarwanda before and after fine-tuning}
\label{fig:3}
\end{figure}

Figure \ref{fig:3} portrays a graphical representation of forgetting and improvement after fine-tuning, for Kinyarwanda.

\section{Conclusion}
This study of benchmarking cross-lingual transfer between Kinyarwanda and Kirundi across both multilingual and monolingual architectures reveals that multilingual models consistently outperform their monolingual counterparts. Multilingual architectures such as mBERT, AfriBERT, and BantuBERT display better accuracy and F1 scores both before and after fine-tuning (FT). In particular, AfriBERT achieves the highest post-FT performance, highlighting its effectiveness in low-resource Bantu languages. Monolingual models like BiGRU, CNN, and Char-CNN, although improving post-FT, lag significantly behind in their initial cross-lingual performance, underscoring the limitations of relying solely on monolingual architectures for cross-lingual tasks. This research affirms the potential of multilingual models in enhancing cross-lingual understanding, particularly in linguistically similar language pairs like Kinyarwanda and Kirundi.

\section{Limitations}
The small size of the training data for both languages limits the model's generalizability to larger datasets or other low-resource Bantu languages. We also did not incorporate continual learning scenarios to mitigate catastrophic forgetting, which could have enhanced performance. While models such as AfriBERT and BantuBERT have shown promising results, their limited pre-training on Bantu languages may impede their ability to fully capture the linguistic intricacies of Kinyarwanda and Kirundi. Furthermore, focusing on just these two languages may restrict the broader applicability of our findings to other Bantu languages.






\section*{Acknowledgments}
\textbf{Funding}: JS. Walia A. Chenat and H. Thangaraj contributed to this work while undertaking a remote collaboration with the Department of Computer Science, University of Pretoria with the Data Science for Social Impact (DSFSI) laboratory with Professor Vukosi Marivate\\\textbf{Open Access}: For open access purposes, the authors have applied a Creative Commons Attribution (CC BY) licence to any Author Accepted Manuscript version arising.\\\textbf{Data Access Statement}: This study involves secondary analyses of the existing datasets, that are described and cited in the text.\\\textbf{Code Access}: \href{https://github.com/karanwxliaa/Cross-lingual-transfer-of-multilingual-models-on-low-resource-African-Languages}{GitHub Repository}.

\bibliography{custom}




\end{document}